\documentclass[conference]{IEEEtran}
\IEEEoverridecommandlockouts
\usepackage{cite}
\usepackage{amsmath,amssymb,amsfonts}
\usepackage{amsfonts}
\usepackage{algorithmic}
\usepackage{mathtools}
\usepackage{graphicx}
\usepackage{textcomp}
\usepackage{xcolor}
\def\BibTeX{{\rm B\kern-.05em{\sc i\kern-.025em b}\kern-.08em
T\kern-.1667em\lower.7ex\hbox{E}\kern-.125emX}}
\newcommand\boldblue[1]{\textcolor{blue}{\textbf{#1}}}
\newcommand{\R}{\mathbb{R}}
\newcommand{\coeff}{\mathrm{coeff}}
\newcommand{\norm}[1]{\left\lVert#1\right\rVert}

\newcommand{\comment}[1]{}

\DeclareMathOperator*{\argmin}{arg\,min}
\DeclareMathOperator*{\argmax}{arg\,max}
\DeclarePairedDelimiter{\ceil}{\lceil}{\rceil}
\begin{document}

\title{A Comparative Study on Structural and Semantic Properties of Sentence Embeddings}

\author{\IEEEauthorblockN{Alexander Kalinowski}
\IEEEauthorblockA{\textit{College of Computing and Informatics} \\
\textit{Drexel University}\\
Philadelphia, PA \\
ajk437@drexel.edu}
\and
\IEEEauthorblockN{Yuan An}
\IEEEauthorblockA{\textit{College of Computing and Informatics} \\
\textit{Drexel University}\\
Philadelphia, PA \\
ya45@drexel.edu}
}

\maketitle

\begin{abstract}
Sentence embeddings encode natural language sentences as low-dimensional dense vectors. A great deal of effort has been put into using sentence embeddings to improve several important natural language processing tasks. Relation extraction is such an NLP task  that aims at identifying structured relations defined in a knowledge base from unstructured text. A promising and more efficient approach would be to embed both the text and structured knowledge in low-dimensional spaces and discover semantic alignments or mappings between them. Although a number of techniques have been proposed in the literature for embedding both sentences and knowledge graphs, little is known about the structural and semantic properties of these embedding spaces in terms of relation extraction. In this paper, we investigate the aforementioned properties by evaluating the extent to which sentences carrying similar senses are embedded in close proximity sub-spaces, and if we can exploit that structure to align sentences to a knowledge graph. 
We propose a set of experiments using a widely-used large-scale data set for relation extraction and focusing on a set of key sentence embedding methods. 
We additionally provide the code for reproducing these experiments at https://github.com/akalino/semantic-structural-sentences.
These embedding methods cover a wide variety of
techniques ranging from simple word embedding combination to transformer-based BERT-style model. Our experimental results show that different embedding spaces have different degrees of strength for the structural and semantic properties. These results provide useful information for developing embedding-based relation extraction methods.

\comment{
Both text documents in unstructured forms and structured sources, such as knowledge graphs, contain semantically valuable information.
Being able to link these diverse data sources is a problem critical to both knowledge integration and information extraction.
A critical choice in a knowledge integration or extraction pipeline is how to represent the data in such a way that it captures the underlying semantics but is also compact and computationally efficient.
}
\comment{
We approach the problem by embedding both the text and graph data in distinct low-dimensional spaces and learning a simple alignment between these spaces to establish correspondences.
We experiment with a variety of embedding techniques and develop an intuition as to what properties of the resulting spaces are desirable as they relate to the structure of the spaces and the semantics they capture.
}
\end{abstract}

\begin{IEEEkeywords}
Sentence Embedding, Knowledge Graph Embedding, Embedding Alignment, Embedding Space Analysis, Relation Extraction
\end{IEEEkeywords}

\section{Introduction}
Embedding spaces have become a staple in representation learning ever since their heralded application to natural language in a technique called word2vec~\cite{word2vec}, and have replaced traditional machine learning features as easy-to-build, high-quality representations of the source objects.
Extending these approaches to larger linguistic units, such as sentences composed from the infinitely expansive combinatorics of language, has generated a plethora of research and techniques.
While sentence embeddings have proven to be useful in applications such as question answering, machine translation, semantic search and text classification, little has been done to explore their structural and semantic properties.
Sentence embeddings as low-dimensional vector representations that reflect certain semantic properties have the potential for applications in efficient and effective relation extraction.
More specifically, we are interested in evaluating the extent to which sentences carrying similar senses are embedded in close proximity sub-spaces, and if we can exploit that structure to align sentences to a knowledge graph.

As an example, if we know that sentence $A$: \texttt{`Dr. Seuss first published The Cat in the Hat in 1957.'} expresses the relation `$author\_of$' and has a sentence embedding representation $v_1 \in \mathbb{R}^n$, we would expect that the semantically similar sentence $B$: \texttt{`Published in 1869, Tolstoy's War and Peace is widely regarded as his magnum opus.'} would have an embedding $v_2 \in \mathbb{R}^n$ such that $v_1$ and $v_2$ share some common features in a subspace of the $n$-dimensional space, and projecting these vectors to this lower dimensional subspace will cause them to have a high cosine similarity.
Thus, their similarity score could be used to infer the `$author\_of$' label for sentence $B$, assisting in bootstrapping the relational labeling process for new text data.

\begin{figure}[htbp]
\centerline{\includegraphics[width=0.5\textwidth]{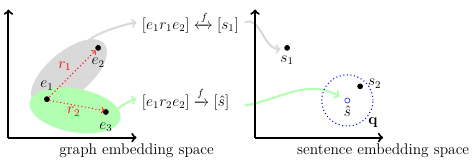}}
\caption{An example of learning a mapping between a knowledge graph embedding space and a sentence embedding space. The learned map $f$ projects the triple embedding representation to the known sentence representation, but should also generalize to the unknown sentence $\hat{s}$. The dotted blue circle represents the nearest neighborhood, contailing the correct sentence $s_2$.}
\label{fig:alignment}
\end{figure}

Furthermore, if we can establish correspondences between sentence representations and structured data source embeddings, such as knowledge graph embeddings, via some map function, we can further automate new data labeling and generalize to detecting the presence of relations in text.
In keeping with the previous example, if we had a general function $f$ such that $f(A) \approx t$, where $t$ is a triple representation of $\langle Dr. Seuss, author\_of , The Cat in the Hat \rangle$, we would like $f$ to also carry sentence $B$ to the triple $\langle Tolstoy, author\_of, War and Peace \rangle$.
This approach would allow for the automated loading of new facts to the graph or development of human-in-the-loop fact verification systems. Fig.~\ref{fig:alignment} illustrates the general situation of applying an alignment between a sentence embedding and a knowledge embedding space, and how to discover similar semantics in nearest neighborhood.

Our aim is to investigate a sample of sentence embedding techniques, ranging from simple word-aggregation methods through advanced Transformer-based neural architectures, evaluating each to measure how useful they are in building well-structured embedding spaces that capture semantics.
For example, methods such as SentBERT~\cite{sentencebert}, InferSent~\cite{infersent}, SkipThought~\cite{skipthought} and many others are explored in our experiments, as well as several others detailed in Section~\ref{sentence-models}.
In this paper, we undertake experiments to evaluate the degree to which sentence embedding methods exhibit structural regularities that can be exploited for further downstream tasks, evaluated through clusterability metrics and mappings to structured semantic data sources.

Clusterability is an important feature of these embedding spaces, especially when using the cosine similarity as a metric for determining the similarity between two objects.
Clusterability can be defined as the degree to which an underlying dataset has the tendency to cluster, or exhibit a grouping structure~\cite{clusterability}.
Spaces with no clusterability, i.e. those that are uniformly randomly generated, will have no notion of similarity between like-objects, while those that capture this notion of similarity will cluster and exhibit structural regularities.
We explore and evaluate clusterability metrics for a variety of sentence embedding methods and show that while all methods perform better than random embeddings, the degree to which these methods are successful in capturing similarities is quite varied.
In addition to clusterability, we additionally desire sentence embeddings to carry semantic properties.
To evaluate the degree to which semantics are captured, we leverage a supervised dataset of sentences and their counterparts in a structured domain, specifically a knowledge graph.
By generating knowledge graph embeddings to encode the semantics of the graph, namely the relationships between entities, we evaluate the extent to which a simple (linear) map can project sentence embeddings to knowledge graph embeddings.
This type of linear mapping, while simplistic, is a white box method that can be further probed to discover which regularities are being carried from one embedding space to another.
Our intent is not to develop a state-of-the-art model for the highest precision mapping between spaces, but simply to evaluate the degree to which different sentence embedding methods can be mapped via linear methods to their semantically grounded counterparts.

The motivation behind this line of experimentation is to use embedding methodologies as a channel through which semantic mappings from unstructured data to structured sources can be generated.
While there is much research on aligning structured sources to other structured sources, such as in ontology mapping~\cite{ontology-matching} or database schema mapping~\cite{schemaMatchingMapping}, there is little research on mapping unstructured sources to their structured counterparts.
It is our hypothesis that the low-dimensional representations built by sentence embedding algorithms are not all created equally – some reflect the semantics of the original data better than others, and this reflection is important to probe, especially as it relates to other information extraction or data integration tasks.
We hope that our experiments motivate further research in this field and provide a starting point for which sentence embedding methods are most effective for this task.
The contributions of this article are as follows.
We present a comparative analysis of several sentence embedding algorithms, evaluating their ability to build a structured vector space and capture the underlying semantics of the knowledge expressed in these sentences.
In addition to this evaluation, we introduce a novel alignment task between structured and unstructured embedding spaces.
This task serves a dual purpose: we use it to evaluate the efficacy of sentence embedding algorithms as well as demonstrate their applicability to relation extraction tasks.
We proceed by describing our experimental design, results and key takeaways, followed by a review of the relevant technical literature.

The rest of the paper is structured as follows.
We provide a motivating example in Section~\ref{example}.
We discuss the relevant technologies for sentence embeddings, knowledge graph embeddings and alignment techniques in Section~\ref{background}.
 Section~\ref{design} discusses our dataset and experimental designs for assessing structure and semantics of embedding spaces.
 Our main results and discussions are presented in Section~\ref{results}, followed by a comparison to related work in Section~\ref{related}.
 We conclude and discuss future research avenues in Section~\ref{conclusion}.

\section{Motivating Example}\label{example}

Relation extraction~\cite{cotype} is a key NLP task that aims at identifying structured relations defined in a knowledge base from unstructured text. Supervised solutions require a large amount of labeled sentences for training relation recognition models.
Adding additional metadata to sentences, such as labelling any named entities or the relations between those entities, is still largely a task that requires time-consuming human annotation.
Seeking to reduce the requires level of human annotators,the work in ~\cite{Mintz_distantsupervision} introduced the idea of distant supervision.
Distant supervision relies on an external data source, namely a knowledge graph, to supply potential labels for a corpus.
Using these `distant' sources reduces the need for fully supervised training sets for relation extraction, yet are known to be noisy and error prone.
Distant supervision assumes that mentions of two entities in a knowledge graph triple indicates the existence of a relation between them, however this is not always the case.
Consider the sentence \texttt{`Thomas Edison was reliant on the availability of a phonograph in his development and subsequent invention of the Kinetograph'}.
This sentence does not express the relation $\langle Thomas Edison, inventor\_of , phonograph \rangle$, but that fact is likely to be contained in our knowledge graph, thus while true, applying the $inventor\_of$ label in the context of the phonograph is incorrect.
To assist in de-noising labels from distantly supervised datasets, we believe that capturing additional semantics via embedding techniques can play an important role, and exploring the degree to which these techniques can help is a goal of this work.
In order to apply these techniques, it is critical to understand and assess the extent to which they capture the structure and semantics of our underlying data.

\section{Background}\label{background}

There has been a wealth of study around techniques for embedding objects, such as images, natural language and knowledge graphs.
These representation learning algorithms, or embeddings, have become a de-facto approach for generating dense and compact feature sets, eliminating the need for tedious human engineering of features at the onset of every new task.
The success of these techniques is not only related to their proven accuracies in downstream tasks, but their ability to train without supervision, allowing them to scale to massive datasets.
While these representations are simple to generate and use as unsupervised features to feed machine learning systems, leading to state-of-the-art accuracies, studies around their properties and probes into exactly what kinds of information they are capturing have been limited.
Here, we focus on comparing multiple sentence representation approaches, evaluating the degree to which they capture the semantics of a sentence, with an eye toward evaluating how relationships between entities are represented in these low-dimensional vector spaces. In the following sub-section, we cover a variety of sentence embedding methods ranging from simple word embedding composition to sophisticated deep neural networks including transformer-based BERT-like model.

\subsection{Sentence Embedding Methods}\label{sentence-models}
Building on the successes of word embedding model~\cite{glove}, a logical next step is to use words as atomic units that compose themselves into sentences.
In this shift, moving from a discrete world where words typically represent a handful of semantic units to a continuous representation in a sentence or document, where words can be combined in infinitely many ways, represents a significant challenge.
Beginning with sub-word embeddings, an efficient sentence classification model is established in~\cite{fasttext} by representing a sentence as the average of its component word representations.
Taking the average or sum of static vectors is a common approach to move from word representations to sentence representations, as discussed in~\cite{sentence-analogies}, and can often beat more advanced models while retaining a level of simplicity.
In our work, we refer to this approach as \textbf{GloVe-mean}.

Even though these representations provide successful baselines, they throw out an important element of data in moving from words to sentences: word ordering.
To address this, ~\cite{discretecosine} propose utilizing a discrete cosine transformation.
By stacking the individual word vectors $w_1, \dots , w_N$ into a matrix, the discrete cosine transformation can be applied column-wise: for a given column $c_1, \ldots , c_N$, a sequence of coefficients can be calculated as
\begin{center}
$$coef[0] = \sqrt{\frac{1}{N}}\sum_{n=0}^N c_n$$
\end{center}
and
\begin{center}
$$coef[k] = \sqrt{\frac{2}{N}}\sum_{n=0}^N c_n \cos \frac{\pi}{N} (n + \frac{1}{2})k$$
\end{center}
The choice of $k$ typically ranges from 0 to 6, where a $k$ of zero is essentially the same as vector averaging, while higher orders of $k$ account for greater impacts of word sequencing.
In our work, we refer to this method as \textbf{GloVe-DCT}.

Alternative approaches abandon static word vectors and focus on the sequence of words in the sentence as the starting point.
To acommodate sequential data, recurrent neural networks (RNNs) dominated the field for a period of time.
Recurrent neural network architectures provide an added benefit that they can theoretically process sequences of variable length (in practice, this is up to some max length where other sequences are padded with a special token), allowing them to train on corpora with long and short sentences.
Another key advantage of RNNs is their ability to share parameters over time, where signal from a prior word carries forward to the next word, and so on.
This benefit of carrying information forward through the network has a downside of making them hard to train, as gradients need to be propogated backward through time; this has caused them to fall out of favor.

One of the first such RNNs trained for sentence encoding was the \textbf{Skip-Thought} model~\cite{skipthought}.
Rather than use word-context windows, the Skip-Thought model generates an encoding for a center sentence and uses that encoding to predict $k$ sentences to the left and right, where $k$ is again the window size as in the skip-gram model.
To accomplish this, the model leverages an encoder-decoder architecture, where each encoder step takes the sequence of words in the sentence and represents them as a hidden state, which is then encoded through the RNN.
Decoding then takes place in two steps, one for predicting the next sentence and one for the prior sentence, each of which generates a hidden state through time that can be used to calculate the probabilities of each word in the sequence, with the following objective function
\begin{center}
$$\sum_t \log P(w_{i+1}^t| w_{i+1}^{<t}, h_i) + \sum_t \log P(w_{i-1}^t| w_{i-1}^{<t}, h_i)$$
\end{center}

An extension of Skip-Thought is the \textbf{Quick-Thought} model~\cite{quickthought}.
The authors note that the objective function used in Quick-Thought is focused on re-creating the surface forms of each sentence given its dependence on the individual words represented.
Specifically, the authors claim ``there are numerous ways of expressing an idea in the form of a sentence. The ideal semantic representation is insensitive to the form in which meaning is expressed''~\cite{quickthought}.
The objective function of Quick-Thought is thus changed to focus only on sentence representations, using a discriminative function to predict a correct center sentence given a window of context sentences.

The authors of~\cite{infersent} continue on the quest to capture sentence semantics with the InferSent model.
InferSent uses a supervised learning paradigm, where sentences in the training set are fed into a three-way classifier, predicting the degree of their similarity (similar, not similar, neutral).
Coupled with a bi-directional LSTM model, the InferSent model can be pre-trained on natural language inference (NLI) tasks such as sentence semantic similarity and later used for inference or fine-tuning on other tasks.
InferSent comes in two flavors, a V1 model using the pre-trained GloVe vectors and a V2 model using pre-trained FastText vectors. In our work, we refer to them as \textbf{InferSentV1} and \textbf{InferSentV2}.

Similar to the InferSent model, \textbf{Sentence-BERT} uses supervised sentence pairs to learn a similarity function~\cite{sentencebert}.
The input sentence embeddings used for the three-way similarity classifier are generated from a pre-trained BERT model.
Contextual models such as BERT provide an encoding of each positional word in an input sentence as their output, thus it is necessary to aggregate these contextualized representations into a single static sentence representation.
As with word embedding models, this aggregation can be a sum or average of the representations at a particular layer of the language model, typically the top layer or final layer.
An alternative option is to provide the model a special classification token ``[CLS]'' that has been pre-trained to compress the contextual representations into one layer, which can then be fed to a non-linear unit.
Sentence-BERT also adapts the classification task to one of regression, where cosine similarity scores are used to score the degree of similarity between sentences.

Continuing on the path of larger, deeper architectures powered my more data, ~\cite{laser} train a Bi-directional LSTM model on a massive scale, multilingual corpus to generate sentence embeddings.
Using parallel sentences accross 93 input languages, the authors are able to focus on mapping semantically similar sentences to close areas of the embedding space, allowing the model to focus more on meaning and less on syntactic features.
Each layer of the LASER model is 512 dimensional, with an output concatenation of both the forward and backward representation generating a final sentence representation of dimension 1024.
The model outperforms BERT-like architectures for a variety of tasks, including cross-lingual natural language inference, a task focused on detecting sentence similarities. In this work, we refer to it as \textbf{LASER}.

An alternate approach that eschews more advanced neural architectures was introduced in~\cite{gem}.
Rather than building new representations, the authors introduce the geometric embedding algorithm (GEM) that focuses on tuning existing pre-trained language representations.
By analyzing the geometry of the subspace generated by building a matrix $A = d \times n$ for the $n$ words in a given sentence, the authors are able to build a new orthogonal basis vector that captures the general semantics of the words and their contexts.
Applying a sliding-window QR factorization to the matrix $A$ and re-weighting by three metrics: a words novelty, significance and uniqueness, a new semantically motivated sentence representation $c_s$ is generated.
Not only is the approach computationally efficient, but it allows for the re-purposing of pre-trained word vectors, such as those from GloVe, allowing for new representations to be generated using no new training data.
In our work, we use the \textbf{GEM $+$ GloVe} version to generate sentence representations.

\subsection{Knowledge Graph Embedding Methods}

Building on the success of embedding-based methods in natural language processing, these techniques have spilled into the domain of knowledge graphs.
Their main motivating uses are in the task of statistical representation learning, where the larger graph is compressed into a low-dimensional representation that can be used by reasoning systems, and knowledge base completion (KBC), where embeddings of existing facts can be utilized to predict new relationships between entities in the graph.
Approaches in this area can be classified into three main categories: translation-based models, semantic matching models and graph structure models.
Our aim is to introduce models leveraged in the alignment literature; for more comprehensive introductions, see~\cite{nickel-survey} and~\cite{ke-survey}.

Let $G = (E, R)$ be a knowledge graph consisting of a set of entities $E$ and relations $R$, each element of which may have an entity or relation type.
From this graph, we can construct the set of known facts, represented as triples $\langle h, r, t \rangle$ with $h,t \in E$ and $r \in R$.
The intuition behind translation-based models is we wish to have low-dimensional, dense representations of $h, r, t$ such that $h + r \approx t$.
Model choices then depend on which space or spaces the entities and relations are embedded in as well as the scoring function used to help the model learn to differentiate between true triples from the graph and noise triples that do not reflect real-world facts.
TransE~\cite{transe} is the simplest of these models.
It embeds both the entities and relations in the same low-dimensional vector space and uses a simple distance function defined by
\begin{center}
$$f_r(h,t) = -\norm{h + r - t}_{1/2}$$
\end{center}

While this model is simple, it struggles to properly encode one to many triples, where a single relation may hold between a head entity and several tail entities.
Resolutions to this are addressed in TransH~\cite{transh}, where each relation is assigned its own hyperplane.
Similarly, TransR~\cite{transr} lets each relation have its own distinct embedding space, greatly expanding the parameter space of the model but increasing the capability of learning relation-specific translations.
An entire family of translation-based models exists, adding various complexities and constraints to the embedding spaces and operations used to best recover the relations described in the original graph.
We leave the exploration of these and more complex methods~/cite{complex, conve} to future work.

\subsection{Alignment Techniques}
Abstractly, learning a mapping function between two vector spaces is a well studied problem\cite{survey-cross-lingual}. Let $\mathcal{D}_1$ and $\mathcal{D}_2$ be two data sets, originating from either similar (as is the case for two language corpora from different languages, or from one graph to another~\cite{kg-align-survey}) or different (as is the case for a set of images and a language corpus) domains.
Let the functions $f_1\colon \mathcal{D}_1 \rightarrow \R^n$ and $f_2\colon \mathcal{D}_2 \rightarrow \R^m$ represent two mappings from the original data sets to their respective real-valued embedding spaces.
Typically, $n$ and $m$ are of much lower dimension than the original cardinalities of $\mathcal{D}_1$ and $\mathcal{D}_2$, and therefore $f_1$ and $f_2$ can be thought of techniques to compress the original data sets whilst maintaining their defining geometric characteristics, including a notion of semantic similarity.
These similarities are measured in the lower dimensional vector spaces through techniques such as, but not limited to, Euclidean distance or cosine similarity.

Let us assume that these semantic similarities are preserved by the functions $f_1$ and $f_2$.
If there exists a correspondence between elements $x \in \mathcal{D}_1$ and $y \in \mathcal{D}_2$, then the problem of aligning their respective embedding spaces seeks to find a map $A\colon \R^n \rightarrow \R^m$ such that $A(f_1(x)) \approx f_2(y)$.

More generally, these methods seek to detect and exploit \textit{invariances} between pairs of low-dimensional embedding spaces.
The degree to which these invariances can be captured dictates how much training data is required to learn a reliable alignment model.
In the case where the underlying geometric structures of both embedding spaces are perfectly invariant, up to a rotation of the space, simple maps may be learned in a highly unsupervised way.
On the flip-side of the coin, methods which do not generate well structured embedding spaces may require more training data in order to learn alignments.
Critically, the problem of learning an alignment map $A$ is also tied to the choice of good embedding functions $f_1$ and $f_2$, and careful coordination between all three choices is required for finding an optimal solution.
Our experiments explore the choice of $f_1$ by selecting a variety of sentence embedding methods and evaluating how well they generate invariances to a semantically grounded counterpart space.

Our main motivation in this line of research is the alignment of sentences to knowledge graphs.
The interest in this problem is two-fold.
Firstly, if we are able to align embeddings of triples $\langle h, r, t \rangle$ from the knowledge graph $G$ to sentence embeddings $s$ in a given corpora, these alignments can be used to detect the expression of relationships $r$ in the sentences, aiding in the task of relation extraction.
Secondly, in the opposite direction, if we can align sentences to triples then we can use this technique to assist in the detection of new triples to be added to the knowledge graph from text data, aiding in the automated expansion of a knowledge graph (see Fig.~\ref{fig:alignment}.)
These two problem domains can be viewed as complementary techniques for converting unstructured data in text documents to structured data in a knowledge graph.
Having data in a structured format not only makes it easier for human verification, as in the case of automated fact checking, but also allows for insights into how other machine learning models, such as document classification, are leveraging unstructured data, providing an avenue for explainable AI and model governance.

\section{Approach}\label{design}

To determine if sentence embedding algorithms exhibit desired structural and semantic properties, we probe the embedding spaces they generate to measure their ability to exhibit clusters of like-data as well as their ability to be mapped to semantically structured data sources.
We approach the problem by designing a set of experiments. 
Our experiments are made available in the following repository: https://github.com/akalino/semantic-structural-sentences.
Each experiment is outlined in the following subsections.

\subsection{Clustering Capacity}

Motivated by work on word embedding regularities\cite{exploit-sim}, we wish to probe the sentence embedding spaces generated by a variery of embedding algorithms and measure the degree to which they exhibit an underlying structure.
Our hypothesis is that embedding techniques that exhibit higher degrees of clusterability are able to capture more of the syntactic and semantic regularities of the input text data sources.
A further hypothesis is that sentences expressing similar relationships should cluster together in the embedding space, creating a structure that will be useful for downstream tasks.

To formalize this notion, we introduce a definition of clusterability, following the work of~\cite{cluster-theory}.
For some dataset $X \subseteq \mathbb{R}^n$, a description of the clusterability of $X$ is a function $c: X \rightarrow v$ where $v \in \mathbb{R}$ is a real value.
Here, $v$ is a measure of how strong a clustering presence is in the underlying set $X$.

Our process for measuring embedding space similarity is as follows.
We begin with our input dataset containing sentences from The New York Times (NYT) corpus.
We selected this input data source as it has been utilized in a multitude of relation extraction studies~\cite{RiedelYM10} and is available with entity and relation annotations to be used in our secondary line of probing.
The dataset is available with sentence segmentation available out-of-the-box; we do minimal additional pre-processing (such as splitting tokens on whitespace and casting to lowercase where necessary for input to embedding algorithms) on the sentences to most accurately reflect how the sentences exist in actual unstructured text documents.
Of the available 434,965 sentences, we apply stratified sampling by relation label to ensure that the underlying relation distributions are kept consistent, and reserve approximately 25 percent of sentences for testing in our secondary probe, generating a set of 320,648 sentences.
Every sentence is additionally labeled with the position and label for mentioned entities and the relations between them.
The dataset covers 53 distinct relations that may be present in the sentences; we note that this distribution is not uniform, with many examples of relations such as `$location/country/capital$' and `$business/person/company$' yet few examples of relations such as `$film/film\_festival/location$'.
We account for this skew in distribution when splitting the data set into training, validation and test folds, assuring that each contains a similar representation of relations.
These sentences are then fed into a variety of sentence embedding algorithms, each generating a sentence embedding matrix with 320,648 rows and a variable number of features determined by the output dimensionality of the given algorithm.

For algorithm selection, we attempt to cover a wide variety of model classes, including those that use static (such as the GloVe pre-trained vectors) and contextual representations.
Details of our selected sentence representation models can be found in Section~\ref{sentence-models}.
We deviate from the work of~\cite{sentence-analogies} by restricting our evaluation of BERT-based models to only use the SentBERT approach, as it is specifically designed to capture sentence semantics, and drop models such as XLNET and GenSen in favor of the simpler and semantically oriented GEM $+$ GloVe model.
We also include a random embedding of dimensionality 300 to serve as a baseline.
For generating sentence embeddings, we leverage our own home-grown code for GloVe-mean, GloVe-DCT and GEM $+$ GloVe. We leverage the open source implementations of LASER~\footnote{https://github.com/facebookresearch/LASER}, InferSent~\footnote{https://github.com/facebookresearch/InferSent}, QuickThought~\footnote{https://github.com/lajanugen/S2V}, SkipThought~\footnote{https://github.com/ryankiros/skip-thoughts} and SentBERT~\footnote{https://github.com/UKPLab/sentence-transformers}.

To test the clusterability hypothesis, we use the spatial histogram approach to measure the clusterability of each space~\cite{clusterability}.
We perform principal component analysis (PCA) to project down to the two most informative dimensions, split the data into $n$ equal-width bins, count how many points lie in each bin, and compute the empirical joint probability mass function.
The same is then done for 500 sets of uniformly generated points with the same feature dimensionality, and the differences are compared using the Kullback-Leibler (KL) divergence.
We report the mean and standard deviation of each of these experiments as our final estimates of clusterability.
Greater average differences indicate divergence from uniformity, indicating some notion of clustering in the underlying space.
For these experiments, we set the configurable number of bins to 20.

\subsection{Semantic Mapping Capacity}

To test the hypothesis that sentence embeddings and knowledge graph embeddings exhibit similar structures in their low-dimensional spaces, we design an experiment for learning a linear map between these spaces using the same NYT dataset.
For each sentence, we have a labeled head entity, tail entity, and relation between these entities.
These labels have corresponding RDF MIDs that allow the corresponding elements of the Freebase knowledge graph to be identified.
Thus, for each labeled example, we have a sentence $s_i$ accompanied with a triple from the graph $\langle h_i, r_i, t_i \rangle$.
As discussed in the clustering capacity probe, we can generate a low-dimensional sentence representations for each of the sentence embedding algorithms selected.
Given that we are interested in developing a baseline to evaluate the efficacy of each sentence representation algorithm, we fix the knowledge graph embeddings by using only the TransE approach as a baseline.
We run TransE over the graph using an embedding dimension of 300, and create a final representation of the triple by concatenating the resulting vectors into a single vector, i.e. $[h:r:t]$, leading to a final representation that is of dimension 900.
Since TransE generates a fixed vector for each of the 53 relations in the graph, we would expect that triples $t_i$ and $t_j$ would occupy similar areas of the embedding space when expressing the same relation $r$.
This property, as enforced by our concatenated representation, gives the embedding space a topological structure we hope to exploit when building the mapping to sentences.
This also allows us to evaluate how well the sentence representations capture the relational semantics contained within the sentences themselves.
Should the embeddings of sentences with similar relations occupy like-areas of the embedding space, building a map to exploit this structure will lead to higher accuracy in the information extraction task.

Suppose we have selected two embedding algorithms $f_1$ and $f_2$ for word embeddings and knowledge graph embeddings, respectively.
Given batches of paired sentences and triples $(s_i, t_i)$, we wish to learn a transformation matrix $W$ to minimize the objective function
\begin{center}
$$\sum_{i=1}^{n} \norm{Wf_1(s_i) - f_2(t_i)}^2$$
\end{center}
Of the 320,648 sentences we reserved for training, we further reserve 114,317, or 35 percent, for model validation based on the same stratified sampling technique by relation.
This larger validation set allows us to prevent our simple linear model from overfitting and to stop training when loss on the validation set begins to increase.
For evaluation on the remaining test set, we generate sentence embeddings for each sentence using each algorithm.
These test embedding matrices are then multiplied by our learned linear map weight matrix $W$ to project the sentence representations into the knowledge graph represention space.
We use all triples available in Freebase as projected by TransE and our concatenation operation to build out the entire knowledge graph space.
From this space, we build an approximate nearest neighbor lookup tree using the Annoy~\footnote{https://github.com/spotify/annoy} approach, with cosine as the similarity metric.
Each new mapped test sentence is then looked up in the nearest neighbors tree based on its highest ranked cosine similarities.
Following in the tradition of knowledge graph embedding literature~\cite{nickel-survey}, we evaluate these results for the Hits@5 and Hits@10 metrics.
For the top $k$ predictions $\hat{v}_{i,k}^{(j)}$ for value $v_i^{(j)}$ we define
\begin{center}
$$Hits@k = \frac{1}{N} \sum{\mathbf{1} (v_i^{(j)} \in \hat{v}_{i,k}^{(j)})}$$
\end{center}
By expanding the query results to the top 5 and top 10 closest matches, we can account for instances where the cosine similarities may result in ties and further analyze the nearest neighborhoods of each projected vector.

For building the TransE embeddings, we use our own implementation in PyTorch. All alignment models and data loaders are built in PyTorch as well.
To level the playing field and focus on the influence of sentence representations in this task, we fix all mapping hyperparameters as follows.
We apply the Adam optimization algorithm with an initial learning rate of 0.001 for 100 epochs.
Each batch is of size 512, and we apply early stopping with a patience of 10 epochs based on the validation set loss.
We further set the optimizer beta to 0.01 and allow for learning rate changes to occur at each epoch.
On both the sentence embedding and knowledge graph embedding features, we apply a normalization scheme consistent with that of~\cite{multistep-linear}.
We also apply an orthogonality constraint to the learned mapping matrix, re-adjusted after each batch.

\section{Results and Analysis}\label{results}

We split our evaluation into two sections: the first on measuring the clusterability of the sentence representations and second on the semantic mapping task.

\subsection{Clusterability - Results}

In addition to efficiently aligning the source and target embedding spaces for information extraction, we are also interested in generating semantically meaningful sentence representations that can be used for relation identification.
To this extent, we are interested in measuring the amount of relational information contained in each distinct sentence embedding approach.
One such way of measuring this information is to measure the clusterability of each embedding space.

\begin{table}[htbp]
\caption{\label{tab:clust-res}Results of clusterability experiments, measured by the KL divergence mean and standard deviation for each spatial histogram.}
\begin{center}
\begin{tabular}{| l | c | c |}
\hline
Sentence Model & Spatial $\mu$ & Spatial $\sigma$ \\ \hline
Random & 0.0018 & 0.0001 \\ \hline
GloVe-mean & 2.1680 & 0.0060 \\ \hline
GloVe-DCT & 1.2412 & 0.0034 \\ \hline
GEM $+$ GloVe & \textbf{4.9716} & 0.0185 \\ \hline
SkipThought & 2.9001 & 0.0075 \\ \hline
QuickThought & 1.2082 & 0.0034 \\ \hline
LASER & 2.0528 & 0.0055 \\ \hline
InferSentV1 & 2.0010 & 0.0055 \\ \hline
InferSentV2 & 2.3067 & 0.0060 \\ \hline
SentBERT & 1.2141 & 0.0033 \\ \hline
\end{tabular}
\end{center}
\end{table}

These metrics allow us to rank sentence embedding algorithms by those that generate the spaces with the highest and lowest clusterability.

\subsection{Clusterability - Discussion}

The GEM $+$ GloVe algorithm shows best clusterability, likely due to the orthogonal projections of each sentence representation, pushing dissimilar sentences into different areas of the embedding space.
Another key consideration that leads to increased clusterability of the GEM $+$ GloVe approach is the addition of the sentence embedding `de-noising' process.
Several studies, namely~\cite{tough-to-beat}, have noted that individual word or sentence vectors all contain a shared, common vector.
This vector contributes to a shared similarity for all representations in a dataset, creating confusion when evaluating metrics such as cosine similarity.
By identifying and removing these like-components, increased distinction between sentence vectors can be achieved~\cite{but-the-top, frequency-agnostic}.
Similar patterns have been noted in~\cite{how-contextual}, where BERT and other deep neural architecture models have been shown to exhibit high levels of anisotropy: all representation vectors have the tendency to occupy a narrow cone in the low-dimensional space.
Eliminating these principal components reduces the tendency for all representations to ``bunch up'' in one area of the space and leads to better performance when measuring clusterability.
We also note that SkipThought performs well in terms of clusterability; however, this is likely due to the size of the embedding dimension (4800) as compared to more compact embeddings.
The other algorithms fall into two main groups: those with a KL divergence close to 2 (GloVe-mean, LASER, InferSentV1, InferSentV2) and those with KL divergence closer to 1.2 (GloVe-DCT, QuickThought, SentBERT).
Given that GloVe-mean is a very simple method of aggregating word vectors into a sentence representation, it is quite surprising that more advanced methods such as LASER and InferSent perform on par when measuring clusterability.
This further demonstrates the main thesis of~\cite{tough-to-beat}, indicating that simplistic methods and insightful geometric post-processing are the keys to generating semantically meaningful and highly clusterable representations.

\begin{figure}[htbp]
\centerline{\includegraphics[width=0.53\textwidth]{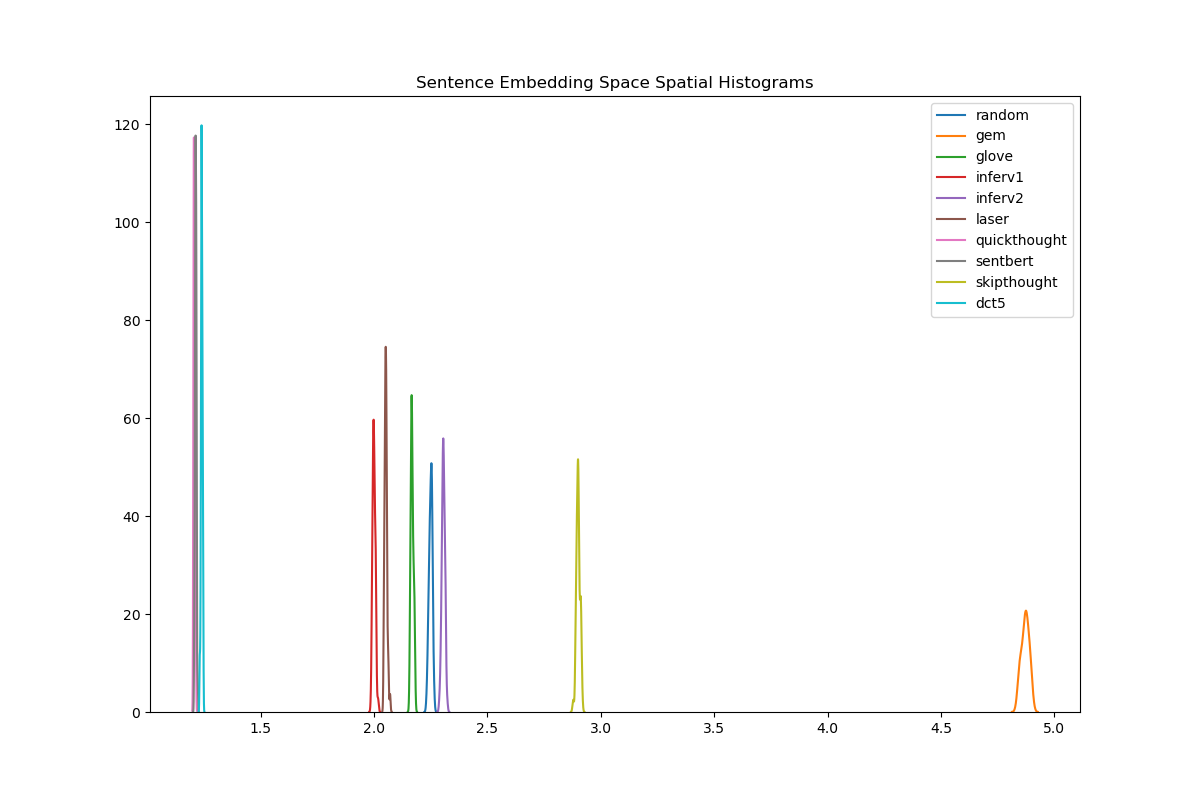}}
\caption{Spatial histograms of sentence embedding clusterability as computed by the mean KL-divergence from a uniformly generated dataset. Visually, we can see that the majority of our selected algorithms are on par with one another, while GEM $+$ GloVe and SkipThought demonstrate greater levels of clusterability.}
\label{fig}
\end{figure}

In terms of efficiency, the close performance of GloVe-mean to more complex methods suggests that deep, highly-specialized sentence representations may not necessarily capture sentence semantics.
The simpler methods are often easier to explain - word vectors are well understood while sentence representations generated from recurrent-style neural networks present more of a black box.
Also of importance is the focus on particular linguistic units; while GloVe focuses on creating an embedding for each token in the corpus, more advanced techniques apply byte-pair encoding schemes that focus less on individual semantic units and more on syntactic, lexical features.
BPE helps to avoid out-of-vocabulary issues (where an embedding can not be generated for a previously unseen token) but may be reducing the importance of focusing on semantic units.
Contextual-based models also generate distinct word embeddings given the semantic context of a word; however, it has been found that the variability in these representations is quite low~\cite{how-contextual}.
This leads to polysemy not being entirely captured in the embedding space, making it hard to differentiate between word usage when clustering.
While word2vec and GloVe approaches have no direct method for dealing with polysemy, the simple mean or re-weighting of words based on their novel orthogonal components may provide enough context to mitigate this issue when moving to higher-order sentence representations.
While deeper representation models have shown to be highly performant on downstream tasks, that performance is often paired with a high-capacity, expressive model, such as a final few layers for classification or attention mechanisms~\cite{all-you-need}.
We instead are interested in the properties of the embeddings themselves, rather than their usability when paired with other machine learning systems.
We conclude that methods using simple, pre-trained word vectors with additional geometric processing, either through mean aggregation or more advanced zero-training methods such as GEM $+$ GloVe, produce the most desirable sentence representation spaces.

\begin{figure}[htbp]
\centerline{\includegraphics[width=0.53\textwidth]{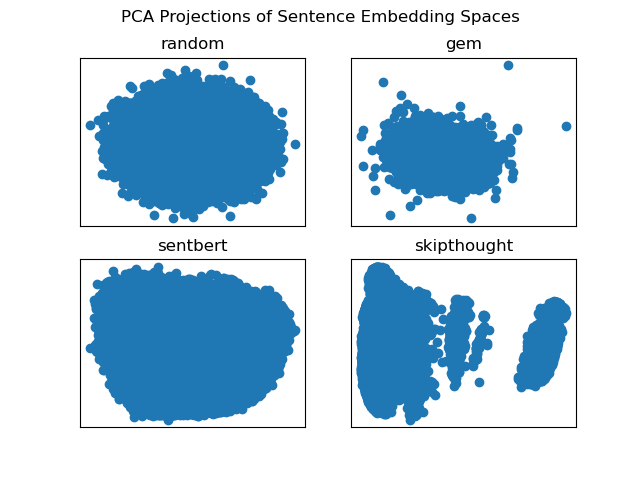}}
\caption{A visualization of PCA projections of selected sentence embedding spaces. Here, we note that both the random space and SentBERT spaces show little indication of structure, while GEM $+$ GloVe and SkipThought demonstrate more variability.}
\label{fig}
\end{figure}

We additionally find that these clusterability metrics do not correlate with sentence embedding dimensionality
($\rho = 0.0031$) because the sentence embedding dimensionality is often times an arbitrary, architecturally driven decision, and not one driven by an attempt to capture the semantics of the sentences themselves.

\subsection{Semantic Mapping - Results}

We present the Hits@5, Hits@10 and average cosine similarity for our selected nine sentence embedding approaches and a randomly generated 300 dimensional sentence space as a baseline below.
Metrics are provided for the test set as described in the experimental design.

\begin{table}[htbp]
\caption{\label{tab:clust-res}Results of semantic mapping experiments, measured by Hits@5 and Hits@10.}
\begin{center}
\begin{tabular}{| l | c | c | c | c |}
\hline
Model & Dim. & Hits@5 & Hits@10 & Avg. Sim. \\ \hline
Random & 300 & 0.0762 & 0.0943 & 0.1195 \\ \hline
GloVe-mean & 300 & 0.1175 & 0.1509 & \textbf{0.5255} \\ \hline
GloVe-DCT & 300 & 0.0249 & 0.0345 & 0.1474 \\ \hline
GEM $+$ GloVe & 300 & \textbf{0.2417} & \textbf{0.3111} & 0.5078 \\ \hline
SkipThought & 4800 & 0.0538 & 0.0665 & 0.2695 \\ \hline
QuickThought & 2048 & 0.0319 & 0.0418 & 0.5062 \\ \hline
LASER & 1024 & 0.0956 & 0.1290 & 0.2479 \\ \hline
InferSentV1 & 4096 & 0.0560 & 0.0824 & 0.3877 \\ \hline
InferSentV2 & 4096 & 0.0598 & 0.0859 & 0.3168 \\ \hline
SentBERT & 768 & 0.1038 & 0.1307 & 0.4806 \\ \hline
\end{tabular}
\end{center}
\end{table}

\subsection{Semantic Mapping - Discussion}

We find that for the purposes of aligning sentence representations with knowledge graph representations, the geometric embedding algorithm (GEM $+$ GloVe) leads to the highest Hits@5 and Hits@10.
This simple processing technique significantly outperforms all other approaches, including those with state-of-the-art neural architectures.
It is our belief that the root cause for this performance gain is due to the means by which the sentence representations are made to fit into semantically similar areas of the embedding space.

Sentences that are semantically expressing the same relationship should occupy a like-subspace of the entire vector space, and are then easier to map to the knowledge graph representations, where our design has forced triple embeddings to occupy the same subspace for each relation.
It is interesting to note that the GloVe-mean, LASER and SentBERT all have similar performance on this task, even with their diversity of approaches and embedding dimensionality.
The advanced architectures of SkipThought, QuickThought and InferSent are consistently outperformed by simpler methods, suggesting that these approaches are either missing out on capturing the semantic meaning of the sentence, or the dimensionality of their embeddings is too large for our simple model to efficiently learn a mapping that generalizes well.
SentBERT, while one of the better performers on the semantic mapping task, shows lower levels of clusterability, highlighting an issue of anisotropy in BERT-like embeddings as discovered in~\cite{how-contextual}.
We also note that the average cosine similarity, computed during evaluation by measuring the pairwise similarities of all testing pairs after the alignment model has been applied, correlates positively ($\rho = 0.514, \rho = 0.503$ for Hits@5 and Hits@10, respectively) with performance.
Average cosine similarities for the more advanced neural models with higher dimensionality tend to be much lower, suggesting that the learned mapping for the alignment task struggles to deal with these high dimensionalities.
Further exploration in either projecting these to a lower dimensional space prior to mapping via a pre-processing method such as PCA, or testing more advanced non-linear models beyond our orthogonally constrained map, may better suite these types of representations.

As expected, the clusterability spatial mean correlates highly with performance ($\rho = 0.8093, \rho = 0.8144$ for Hits@5 and Hits@10, respectively), indicating that spaces that cluster well are easier to build align.
Thus, we conclude that clusterability of embedding spaces is a good proxy metric for evaluating how effective these methods may be for relation extraction tasks.

\section{Related Work}\label{related}

The work most closely related to ours evaluates sentence analogies by comparing a variety of sentence embedding algorithms~\cite{sentence-analogies}.
Their study evaluates the degree to which lexical analogies are reflected in sentence embeddings, particularly those of the form $A$ is to $B$ as $C$ is to $D$, denoted $A:B :: C:D$.
The authors focus explicitly on five classes of semantic relationships, namely common capital cities, all capital cities, currencies, city in state and gender relations.
These five classes are highly common in written language, particularly in news articles, thus representing high frequency instances.
Work has been done to show that frequency of mentions is tied closely to the magnitude of direction of a word in low-dimensional vector space~\cite{frequency-agnostic}, so it is not surprising that this limited set of semantic relationships can be recovered, although the degree to which methods are successful widely varies.
Our work probes a broader class of semantic relationships, and in particular, those that may lack in an abundance of training examples.
Rather than focusing on the syntactic and semantic properties via an analogy task, we focus on evaluating the embedding spaces themselves to determine if they contain useful structural properties that can be applied to further information extraction tasks.
By focusing on the space itself, we can evaluate the degree to which the embedding algorithms capture structural similarities between sentences.
These structural similarities should present themselves through cosine similarities, where $sim(v_1, v_2) \approx 1$ for very semantically similar sentences.

While structural regularities for the task of sentence analogies have been demonstrated in~\cite{sentence-analogies}, it is unclear if those same regularities could be leveraged for semantic mapping tasks.
Our evaluation of similarity focuses instead on the task of semantic mapping, for which we establish correspondences between sentence embeddings and knowledge graph embeddings.
The degree to which these correspondences exist helps to establish how effective semantic mappings created through embedding space alignment can be and builds the case for their application in semantic relation extraction from unstructured text data.

The issue of supervised data set construction is especially important in this area.
Not only are instances of sentences with entity and relation labels hard to collect as human labelers are known to have low precision for the task~\cite{extreme, extreme-mr}, especially in domains such as biomedicine where subject matter experts are required, but the speed at which new entities and relations may be discussed outpaces the development of such datasets, making supervised models stale and unable to generalize quickly.
To combat the problem of data collection, the technique of distant supervision was introduced~\cite{Mintz_distantsupervision}, allowing for entity and relation triples from an ontology or knowledge graph to be used for quick label generation over raw text inputs.
In this paradigm, when two entities related by a relation in the knowledge graph appear in a sentence, that sentence is labeled as being representative of that relation.
This distant supervision assumption and its relaxations allow for quick bootstrapping of labeled datasets, but it is well known that they also introduce a great deal of noise, as is the case when a sentence mentioning two entities expresses a novel relation, causing a false label, and are equally susceptible to missing labels when the surface forms don't match or are ambiguous~\cite{dss2019, noise-reduction}.
While distant supervision techniques still depend largely on linking the ontology and text corpus via surface forms (i.e.\ matching on likely string spans or candidate mentions), we anticipate a growing field of alignment between knowledge graph and language embeddings given the increases in alignment techniques used in the two distinct data domains, a main motivating factor for the undertaking this line of experimentation.

\section{Conclusion and Future Work}\label{conclusion}

In this work, we model the distant supervision for relation extraction task as one of aligning distinct embedding spaces.
Learning an alignment between these low-dimensional spaces allows us to quickly generalize to new instances without the reliance on any surface form matching or rule-based systems.
The approach here is quite generic and adaptable to diverse domains and embedding algorithms.
We find that the approach works best for sentence embedding algorithms that introduce a notion of semantic similarity, where sentences expressing like-relations are easily discoverable in the structure of the low-dimensional space.
We also explore the clusterability of these spaces and find that the GEM $+$ GloVe model is the most clusterable of the available techniques while also providing the best relation extraction performance, tying performance on the task to the geometric structure of the underlying space.
As we aim to assess the feasibility of this alignment process, this paper does not explore all available embedding options nor conduct extensive hyperparameter tuning.
Future work will attempt to substitute the TransE embeddings with other more expressing graph embedding models, with the hope that more advanced techniques will help in the relation extraction task.
We will also continue to explore what gives an embedding space the right semantic structure that can be exploited for ease of alignment.
Additionally, we wish to evaluate the influence on the size of the training data set, looking for methods of alignment that leverage as little supervised data as possible in the hopes of using these techniques to quickly bootstrap other information extraction tasks.

\bibliographystyle{unsrt}
\bibliography{alignment}

\end{document}